\documentclass[10pt,twocolumn,letterpaper]{article}

\usepackage{cvpr}
\usepackage{microtype}
\usepackage{times}
\usepackage{epsfig}
\usepackage{graphicx}
\usepackage{amsmath}
\usepackage{amssymb,bm}
\usepackage{booktabs}
\usepackage{cuted}

\usepackage[pagebackref=true,breaklinks=true,letterpaper=true,hidelinks,bookmarks=false]{hyperref}

\cvprfinalcopy 

\DeclareMathOperator*{\E}{{\mathbb{E}}}

\DeclareMathOperator*{\argmin}{arg\,min}

\newcommand{\secprespace}{\vspace{-2mm}}
\newcommand{\secspace}{\vspace{-1mm}}

\newcommand{\figref}[1]{Fig.~\ref{fig:#1}}
\newcommand{\tabref}[1]{Table~\ref{tab:#1}}
\newcommand{\secref}[1]{Sec.~\ref{sec:#1}}

\newcommand{\appref}[1]{Appendix~\ref{sec:#1}}

\newcommand{\eq}[1]{\eqref{eq:#1}}
\newcommand{\norm}[1]{\left\lVert#1\right\rVert}

\makeatletter
\renewcommand{\paragraph}{%
  \@startsection{paragraph}{4}%
  {\z@}{0.60ex \@plus 1ex \@minus .15ex}{-1em}%
  {\normalfont\normalsize\bfseries}%
}
\makeatother

\def\cvprPaperID{1028} 

\ifcvprfinal\fi
\begin{document}

\title{One Man's Trash is Another Man's Treasure:\\ Resisting Adversarial Examples
by Adversarial Examples}

\author{Chang Xiao\qquad Changxi Zheng\\
Columbia University\\
{\tt\small \{chang, cxz\}@cs.columbia.edu}
}
\maketitle

\begin{abstract}
\vspace{-2mm}
Modern image classification systems are often 
built on deep neural networks,
which suffer from adversarial examples---images with deliberately crafted, imperceptible 
noise to mislead the network's classification.
To defend against adversarial examples, 
a plausible idea is to obfuscate the network's gradient with respect to the
input image. This general idea has inspired a long line of defense methods.
Yet, almost all of them have proven vulnerable.

We revisit this seemingly flawed idea from a radically different
perspective. We embrace the omnipresence of adversarial examples and the
numerical procedure of crafting them, and turn this harmful attacking process 
into a useful defense mechanism. Our defense method is conceptually simple:
before feeding an input image for classification,
transform it by finding an adversarial example on a pretrained external model.
We evaluate our method against a wide range of possible attacks.
On both CIFAR-10 and Tiny ImageNet datasets,
our method is significantly more robust than state-of-the-art methods.
Particularly, in comparison to adversarial training, 
our method offers lower training cost as well as stronger robustness.


\vspace{-2mm}
\end{abstract}

\secprespace
\section{Introduction}
\secspace
Deep neural networks have vastly improved the performance of image
classification systems. Yet they are prone to \emph{adversarial examples}.
Those are natural images with deliberately crafted, imperceptible noise, aiming
to mislead the network's decision
entirely~\cite{biggio2013evasion,szegedy2013intriguing}.  
In numerous applications, from face recognition authorization to autonomous
cars~\cite{sharif2016accessorize,thys2019fooling}, the vulnerability caused by
adversarial examples gives rise to serious security concerns and presses for
efficient defense mechanisms.

The defense, unfortunately, remains grim. Recent
studies~\cite{shafahi2018adversarial,tsipras2018robustness,fawzi2018adversarial} 
suggest that the prevalence of adversarial examples may be an inherent 
property of high-dimensional natural data distributions.
Facing this intrinsic difficulty of eliminating adversarial examples,
a plausible thought is to conceal them---making them hard to find.
Indeed, a long line of works aims to obfuscate the network model's gradient with respect to
its input~\cite{xie2018mitigating, guo2018countering, xiao2019resisting,
buckman2018thermometer,song2018pixeldefend,samangouei2018defensegan}, 
motivated by the fact that the gradient information is essential for crafting adversarial
examples: the gradient indicates how to perturb the input to alter the
network's decision.

\begin{figure}[t]
  \centering
  \vspace{-2mm}
    \includegraphics[width=0.99\columnwidth]{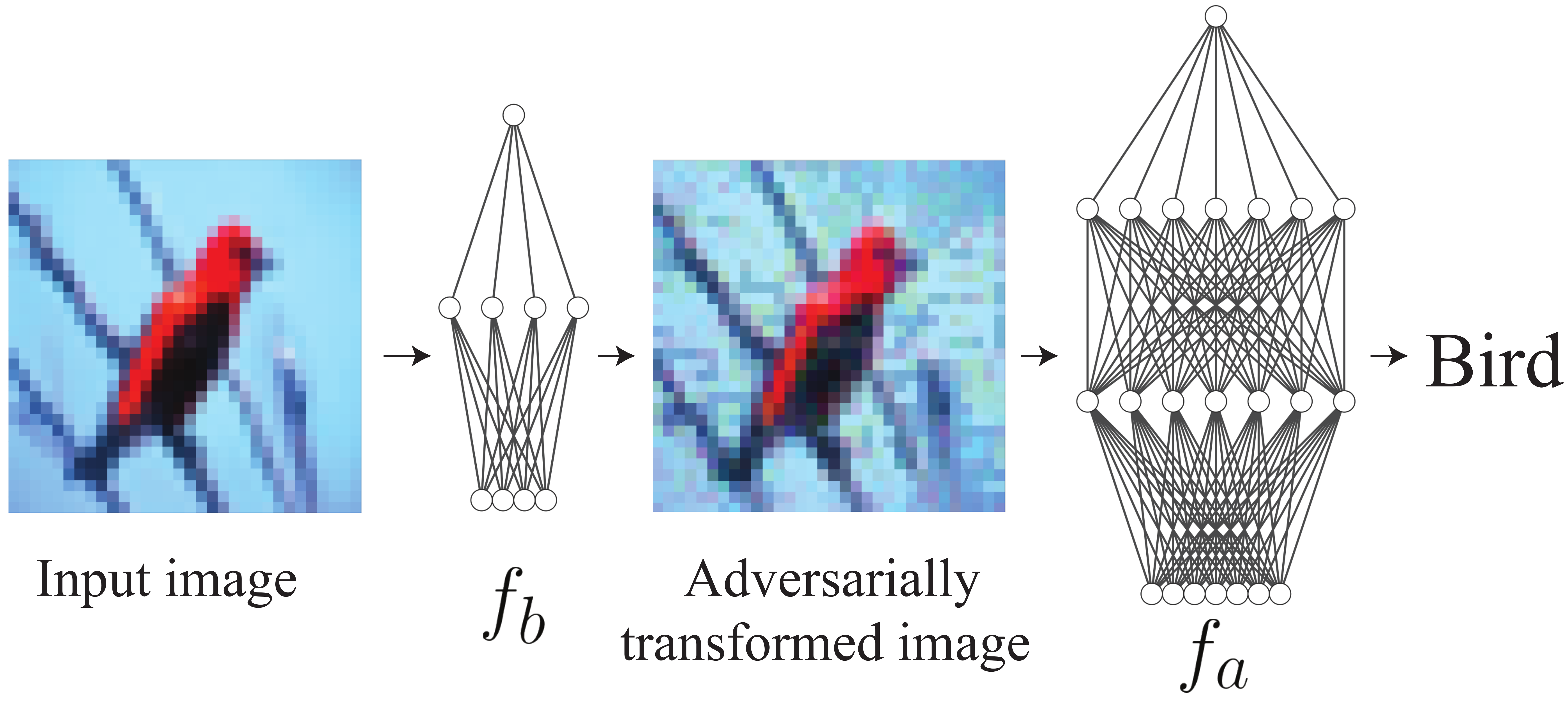}
    \vspace{-1mm}
    \caption{\textbf{A simple and {effective} defense mechanism}.
    Given an input image, our defense method first transforms it
    through the process of crafting adversarial examples on a pre-trained simple model $f_b$,
    deliberately adding strong adversarial noise. The transformed image is 
    then fed into another model $f_a$ for classification.
    The same pipeline is applied in both training and inference.
  }\label{fig:pipeline}
  \vspace{-3mm}
\end{figure}

Yet, almost all these gradient obfuscation based defenses have proven vulnerable. 
In their recent seminal work, Athalye et al.~\cite{athalye2018obfuscated} presented 
a suite of strategies for estimating network gradients
in the presence of gradient obfuscation.
Adversarial examples crafted by their method have successfully fooled many existing defense models, 
some of which even yield 0\% accuracy under their attack.

We revisit the idea of gradient obfuscation but take a radically different approach.
Instead of expelling adversarial examples, we embrace them.
Instead of obstructing the way of finding adversarial examples on a model,
we exploit it to strengthen the robustness of another model.

Our defense is conceptually simple:
before feeding an input image to a classification model,
we transform it through
the process of finding adversarial examples on an external model.
Mathematically, if we use $f(\bm{x})$ to denote the model that classifies 
an input image $\bm{x}$, our defense model is expressed as $f(g(\bm{x}))$,
where $g(\cdot)$ represents the process of finding an adversarial example near $\bm{x}$ on a pre-trained 
external model (see \figref{pipeline}).

The robustness of our defense model $f(g(\bm{x}))$ stems from 
the fundamental difficulties of estimating the gradient of $g(\bm{x})$
with respect to $\bm{x}$.
Finding an adversarial example amounts to searching for a local minimum
on a highly fluctuated objective landscape~\cite{madry2017towards}. 
As a result, $g(\bm{x})$ is not an analytic function, not smooth, not deterministic, but an iterative
procedure with random initialization and non-differentiable operators.
We show that all these traits together constitute a highly robust 
defense mechanism.

We play devil's advocate in attacking our defense model thoroughly.
We examine a wide range of possible attacks, including those having successfully 
circumvented many previous defenses~\cite{athalye2018obfuscated}.
Under these attacks, we compare the \emph{worst-case} robustness of our method 
with state-of-the-art defense methods on both CIFAR-10 and Tiny ImageNet datasets.
Our defense demonstrates superior robustness over those methods.
Particularly, in comparison to models optimized with adversarial training---by
far the most effective defense against white-box attacks---our method offers
simultaneously lower training cost and stronger robustness. 



%
%




\section{Related Work}\label{sec:related}
\secspace
\paragraph{Adversarial attack.}
The seminal work of Biggio et al.~\cite{biggio2013evasion} and Szegedy et
al.~\cite{szegedy2013intriguing} first suggested the existence of adversarial
examples that can mislead deep neural networks.  The latter also used
a constrained L-BFGS to find adversarial examples.  
Goodfellow et al.~\cite{goodfellow2014explaining} later introduced Fast Gradient Sign Method (FGSM)
that generates adversarial examples more efficiently.  
Madry et al.~\cite{madry2017towards} further formalized the problem of
adversarial attacks and proposed Projected Gradient Descent (PGD) method, which
further inspires many subsequent attacking methods~\cite{dong2018boosting, carlini2017towards,moosavi2016deepfool, kurakin2016adversarial}. 
PGD-type methods are considered the strongest attacks based on first-order information, namely
the network's gradient with respect to the input~\cite{madry2017towards}.  
To compute the gradients, the adversary must have full access to the network structure and parameters.
This scenario is referred as the \textit{white-box} attack.

When the adversary has no knowledge about the model, 
the attack, referred as \emph{black-box} attack, is not as easy as the white-box attack.
By far the most popular black-box attack is the so-called \emph{transfer attack},
which uses adversarial examples generated on a known model (e.g., using PGD) to
attack an unknown model~\cite{papernot2017practical}.
Several methods (e.g.,~\cite{tramer2018ensemble,athalye2018synthesizing, Xie_2019_CVPR,Inkawhich_2019_CVPR}) 
are proposed to improve the 
transferability of the adversarial examples so that the adversarial examples generated on one model
are more likely to fool another model.
Another type of black-box attacking methods is \emph{query-based}~\cite{brendel2018decisionbased,Shi_2019_CVPR,chen2019boundary,alzantot2019genattack,li2019nattack}:
they execute the model many times with different input in order to
learn the behavior of the model and construct adversarial examples.




While our defense is motivated by attacks in white-box scenarios,
we evaluate our method under a wide range of possibilities, including
both white-box and black-box attacks.

\paragraph{Adversarial defense.}
The threat of adversarial examples has motivated active studies of 
defense mechanisms.
By far the most successful defense against white-box attacks is 
\emph{adversarial training}~\cite{madry2017towards, goodfellow2014explaining,sinha2018certifiable},
and a rich set of methods has been proposed to accelerate its training
speed or further improve its robustness~\cite{xie2019feature, zhang2019theoretically,
liao2018defense,goodfellow2014explaining, wang2019bilateral,
shafahi2019adversarial, zhang2019you, Mustafa_2019_ICCV, mao2019metric}.
In comparison to adversarial training, our method offers both
stronger robustness and lower training cost. 

To defend against gradient-based attacks (such as the PGD attack), a natural idea
is to obfuscate (or mask) network gradients~\cite{papernot2017practical, tramer2018ensemble}.
To this end, there exist a long line of works that apply
random transformation to input images~\cite{xie2018mitigating, guo2018countering},
or employ stochastic activation functions~\cite{s2018stochastic} 
and non-differentiable operators in the model~\cite{xiao2019resisting,
buckman2018thermometer,song2018pixeldefend,samangouei2018defensegan}.  


Unfortunately, many of these methods have proven vulnerable by Athalye et al.~\cite{athalye2018obfuscated}, 
who introduced a set of attacking strategies, 
including a method called Backward Pass Differentiable Approximation (BPDA),
to circumvent gradient obfuscation (see further discussion in \secref{input_trans} and \ref{sec:hardness}). 
Since then, a few other gradient obfuscation based defenses have been 
proposed~\cite{Liu_2019_CVPR, Raff_2019_CVPR, Jia_2019_CVPR, Taran_2019_CVPR, liu2018towards}.
But those works either report degraded robustness under BPDA attacks~\cite{Liu_2019_CVPR, Raff_2019_CVPR}
or neglected the evaluation against BPDA attacks~\cite{Jia_2019_CVPR, Taran_2019_CVPR, liu2018towards}.

Thus far, gradient obfuscation is generally considered vulnerable (and at least incomplete)~\cite{athalye2018obfuscated}.
We revisit gradient obfuscation, 
and our defense demonstrates unprecedented robustness against BPDA and other possible attacks.

\section{Defense via Adversarial Transformation}
\secspace

We now present a simple approach to defend against adversarial attacks.
We will first motivate and describe our 
\emph{adversarial transformation} (\secref{input_trans}
and \ref{sec:adv_trans}),
and then provide the rationale of \emph{why} it improves adversarial robustness (\secref{hardness} and \ref{sec:f_b}),
backed by empirical evidence (\secref{eval}).

\subsection{Motivation: Input Transformation}\label{sec:input_trans}
\secspace
An attempt that has been explored in adversarial defense---albeit
unsuccessfully so far---is the defense via input transformation.
Consider a neural network model $f_a$ that classifies the input image $\bm{x}$
(i.e., evaluating $f_a(\bm{x})$).
Instead of feeding $\bm{x}$ into $f_a$ directly,
this defense approach transforms the input image through
an operator $g$ before presenting it to the classification model (i.e., 
evaluating $f_a(g(\bm{x}))$).

The transformation $g$ is applied in both training and inference.
Provided a training dataset $\mathcal{X}$, the network weights
$\bm{\theta}$ are optimized by solving
\begin{equation}\label{eq:training}
    \bm{\theta}^* = \argmin_{\bm{\theta}}\E_{(\bm{x},y)\in\mathcal{X}}
    \left[\ell(f_a(g(\bm{x});\bm{\theta}), y)\right],
\end{equation}
where $\bm{x}$ and $y$ are respectively the image and its corresponding label drawn
from the training dataset, and $\ell$ is the loss function (such as 
cross entropy for classification tasks).
Correspondingly, at inference time, the model predicts
the label of an input image $\bm{x}$ by evaluating $f_a(g(\bm{x}))$.

Input transformation $g(\cdot)$ offers an opportunity to implement 
the idea of gradient obfuscation.
For example, by transforming the input image with certain randomness such as random 
resizing and padding~\cite{xie2018mitigating}, the network gradients become hard 
to estimate. 

Another use of $g(\cdot)$ for defense is to remove the noise (or perturbations) in adversarial examples.
For instance, $g(\cdot)$ has been used to restore a natural image from a potentially adversarial input,
by projecting it on a GAN- or PixelCNN-represented image manifold~\cite{samangouei2018defensegan, song2018pixeldefend}
or regularizing the input image through total variation minimization~\cite{guo2018countering}.

These input-transformation-based defense mechanisms seem plausible.
Yet they are all fragile.
As demonstrated by Athalye et al.~\cite{athalye2018obfuscated}, 
with random input transformation, 
adversarial examples can still be found using 
Expectation over Transformation~\cite{athalye2018synthesizing},
which estimates the network gradient
by taking the average over multiple trials (more details in \secref{hardness}). 
The noise-removal transformation 
is also ineffective. 
One can use Backward Pass Differentiable Approximation~\cite{athalye2018obfuscated} to easily construct effective adversarial examples. 
In short, the current consensus is that input transformation as 
a defense mechanism remains vulnerable. 


We challenge this consensus.
We now present a new input transformation method for gradient obfuscation, 
followed by the explanation of 
why it is able to avoid the shortcomings of prior work and offer stronger adversarial robustness.


\subsection{Adversarial Transformation}\label{sec:adv_trans}
\secspace
Our input transformation operation takes an approach
\emph{opposite} to the intuition behind previous methods~\cite{guo2018countering,song2018pixeldefend,samangouei2018defensegan}.  
In contrast to those aiming to purge
input images of the adversarial noise,
we embrace adversarial noise.
As we will show, our transformation injects noticeably strong adversarial noise into the input image.
This seemingly counter-intuitive operation 
is able to strengthen the network model in training,
making it more robust.

Our transformation operation relies on another network model $f_b$, whose
choice will be discussed later in \secref{f_b}.
The model $f_b$ is pre-trained to perform the same task as $f_a$.
Then, given an input image $\bm{x}$, the transformation operator $g(\cdot)$ is defined 
as the process that finds the adversarial example nearby $\bm{x}$ to fool $f_b$. 
Formally, this process is meant to reach 
a local minimum of the optimization problem,
\begin{equation}\label{eq:attack_b}
    g(\bm{x}) = \argmin_{\bm{x}'\in\Delta_{\bm{x}}}\ell(f_b(\bm{x}'),y_{\text{L}}),
\end{equation}
where $\ell(\cdot)$ is the loss function as used in network training~\eq{training};
and $y_\textrm{L}$ is the adversarial target, setting to be the input $\bm{x}$'s least likely class 
predicted by $f_b$.  The adversarial examples are restricted in
$\Delta_{\bm{x}}$, an $L_\infty$-ball at $\bm{x}$, defined as
$\|\bm{x}'-\bm{x}\|_\infty < \Delta$. The \emph{perturbation range} $\Delta$ is
a hyperparameter.

Transformation $g(\bm{x})$ defined in~\eq{attack_b} can be implemented using
any gradient-based attacking methods (such as
Deepfool~\cite{moosavi2016deepfool} and C\&W~\cite{carlini2017towards}).
We choose to use
the least-likely class projected gradient descent (LL-PGD) method~\cite{kurakin2016adversarial,tramer2018ensemble}.
LL-PGD is an iterative process, 
wherein each iteration updates the adversarial example by the rule,
\begin{equation}\label{eq:at_update}
    \bm{x}'_t =\Pi_{\bm{x}'\in \Delta_{\bm{x}}}\left[\bm{x}'_{t-1} - \epsilon \cdot \text{sgn}(\nabla_{\bm{x}}\ell(f_b(\bm{x}'_{t-1}), y_\textrm{L}))\right].
\end{equation}
Here $\bm{x}'_t$ denotes the adversarial example after $t$ iterations; 
$\text{sgn}(\cdot)$ is the sign function, and
$\Pi_{\bm{x}'\in \Delta_{\bm{x}}}[\cdot]$ projects the image back into the allowed
perturbation ball $\Delta_{\bm{x}}$.
This iterative process starts from a random perturbation of input image
$\bm{x}$, namely $\bm{x}+\bm{\delta}$, where each element (pixel) in $\bm{\delta}$ is
uniformly drawn from $[-\Delta,\Delta]$.
The output $\bm{x}'_N$ (after $N$ iterations) is the transformed
version of $\bm{x}$. In other words, $g(\bm{x}) = \bm{x}'_N$, which we refer as \emph{adversarial transformation}.


After defining the adversarial transformation $g(\cdot)$ based on the pre-trained model $f_b$, we use $g(\cdot)$ to train the model $f_a$
as described in~\eq{training}. At inference time, the label of an image $\bm{x}$ is predicted as 
$f_a(g(\bm{x}))$. Figure~\ref{fig:pipeline} illustrates the pipeline of our method. 

\paragraph{Differences from adversarial training.}
With adversarial transformation, 
our training process 
superficially resembles the adversarial training,
because both training processes need to search for adversarial
examples of the input training data.
But fundamental differences exist. 
In adversarial training, a single model $f_a$ is used for crafting
adversarial examples and evolving itself at each epoch,
whereas our method involves two models:
the model $f_b$ is pre-trained and stays fixed during both the training of $f_a$ and the 
inference using $f_a$.  


The consequence of using a fixed external model $f_b$ for adversarial transformation is substantial.
As we will discuss in \secref{f_b}, $f_b$ can be chosen much simpler than $f_a$.
As a result, crafting the adversarial examples on $f_b$ has lower cost than that on $f_a$,
and thus our training process is faster than adversarial training (see experiments in \secref{comp}).
More remarkably, the adversarial transformation using $f_b$ makes the model $f_a$ much harder to attack, as explained next.

\subsection{Rationale behind Adversarial Transformation}\label{sec:hardness}
\secspace


\begin{figure}[t]
  \centering
    \includegraphics[width=0.85\columnwidth]{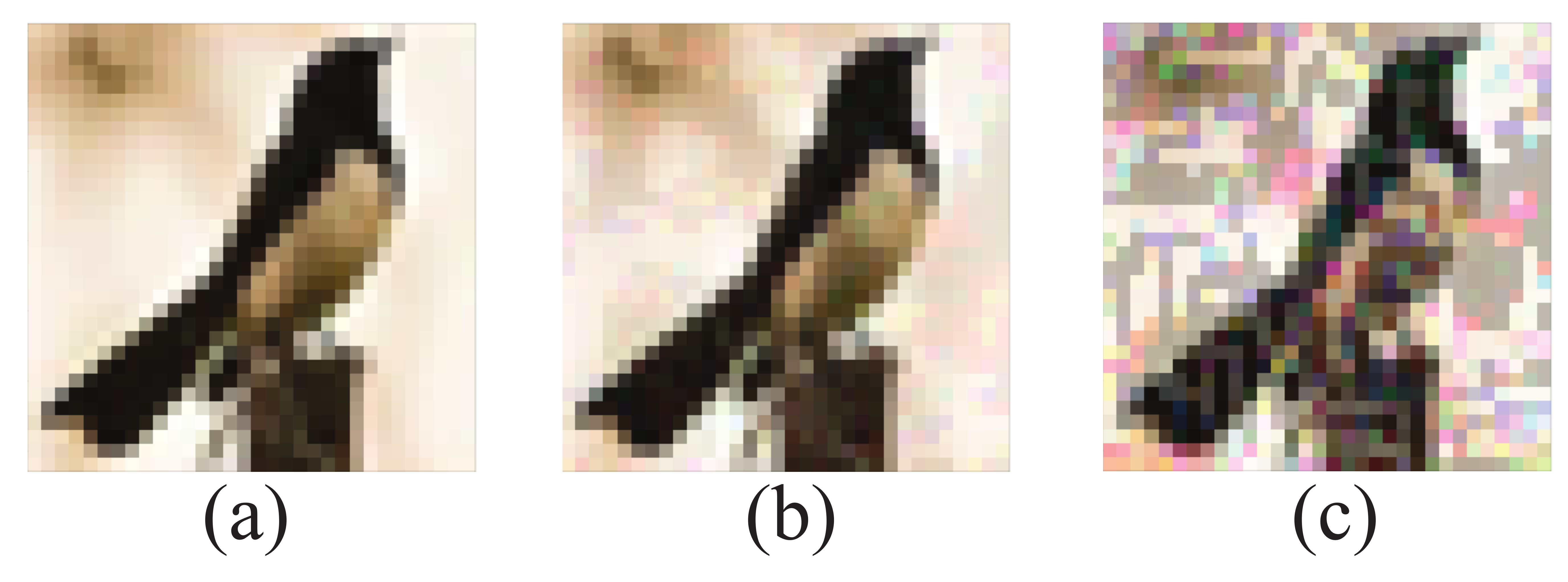}
    \vspace{-2mm}
  \caption{
      To launch a valid attack from an input image (a), the adversarial example (b)
      must be perceptually similar to the original image (e.g., here $\Delta=0.031$).
      Otherwise, it can be easily pinpointed.
 In our method, the transformed image is used for training $f_a$, not attacking. We therefore
 intentionally add much stronger adversarial noise to the input (c) (here $\Delta=0.2$).
 The strong noise helps to strengthen the robustness of $f_a$ and defend 
      against BPDA attacks (see \secref{hardness}). 
  }\label{fig:noise_inject}
    \vspace{-3mm}
\end{figure}
\paragraph{Embracing adversarial noise.}
Given an input image $\bm{x}$, 
our adversarial transformation effectively adds perturbation noise to $\bm{x}$. 
The perturbation range $\Delta$ 
controls how much noise is added. Normally, in adversarial attacks, $\Delta$ is set small
to generate adversarial examples perceptually similar to the input image.
But when we use adversarial attacks (on $f_b$) as a means of input transformation for training $f_a$,
we have the freedom to use a much larger $\Delta$, thereby adding 
noticeably stronger adversarial noise (see \figref{noise_inject}-c).

At training time, the excessively strong adversarial noise forces
the network $f_a$ to learn how to classify robustly. This is because 
perturbations crafted on an external model can approximate the adversarial examples of 
the model under training (an insight inspired the prior work~\cite{tramer2018ensemble}).
This reason, although valid, can not explain how our method is able to 
avoid the deficiencies of prior defense methods. There exist deeper reasons:



\paragraph{Randomness.}
The adversarial noise added by our $g(\bm{x})$ is randomized, since the
update rule~\eq{at_update} always starts from the input image with a random perturbation 
(i.e., $\bm{x}+\bm{\delta}$ with uniformly sampled
$\bm{\delta}_i\sim[-\Delta,\Delta]$).
Randomization is not new;
prior defenses also employ randomized transformations to the input. But they have been
circumvented by Expectation Over Transformation (EOT)~~\cite{athalye2018obfuscated,athalye2018synthesizing}.
EOT attack first estimates the gradient of expected $f(g(\bm{x}))$ with respect to $\bm{x}$
using the relationship $\nabla\E_{\tilde{g}\sim\mathcal{T}}f(\tilde{g}(\bm{x}))
=\E_{\tilde{g}\sim\mathcal{T}}\nabla f(\tilde{g}(\bm{x}))$, 
where $\tilde{g}(\cdot)$ is a deterministic version of
$g(\cdot)$ sampled from the distribution of randomized transformations $\mathcal{T}$.
It then uses the estimated gradients in PGD-type attacks to generate adversarial examples.
Thus, the feasibility of EOT hinges on a reliable estimation of $\nabla
f(\tilde{g}(\bm{x}))$. 
In our method, $\tilde{g}(\bm{x})$ corresponds to solving
the optimization problem~\eq{attack_b} starting from a particular sample $\bm{x}+\bm{\delta}$. 

In what follows, we examine a range of strategies that have been successfully used to 
estimate $\nabla f(\tilde{g}(\bm{x}))$ in prior defense methods (and thus break them), and show 
that our method is robust against all those attacking strategies.

\paragraph{Automatic differentiation.}
By chain rule, the estimation of $\nabla f(\tilde{g}(\bm{x}))$
requires the knowledge of $\tilde{g}(\bm{x})$'s Jacobian (first-order derivatives) $\text{D}\tilde{g}(\bm{x})$.
A straightforward attempt to this end is by unrolling the iterative steps~\eq{at_update}
and using automatic differentiation (AD)~\cite{wengert1964simple} to compute $\text{D}\tilde{g}(\bm{x})$.
Yet, this is infeasible. As shown in~\eq{at_update}, the iterative steps 
involves \emph{non-differentiable} operators including 
$\text{sgn}(\cdot)$ and $\Pi_{\bm{x}'\in \Delta_{\bm{x}}}[\cdot]$.
Thus, directly applying AD
leads to erroneous estimation of $\text{D}\tilde{g}(\bm{x})$, which in turn obstructs the
search for adversarial examples. 
Our early experiments indeed show that virtually no adversarial examples crafted using AD can fool our model. 

\paragraph{Backward Pass Differentiable Approximation (BPDA).}
To circumvent the defense using non-differentiable operators,
Athalye et al.~\cite{athalye2018obfuscated} introduced a strategy called
Backward Pass Differentiable Approximation (BPDA) to estimate the defense model's gradients.
The idea is to replace the non-differentiable operators in $\tilde{g}$ with
differentiable approximations, and estimate the derivatives $\text{D}\tilde{g}(\bm{x})$
in AD by computing the AD's forward pass using the original $\tilde{g}$ and 
computing its backward pass using $\tilde{g}$'s differentiable approximation.
BPDA has succeed in gradient-based attacks (such as PGD and C\&W~\cite{carlini2017towards})
toward many prior defenses, allowing the adversary to
craft efficient adversarial examples.

When applying BPDA to estimate the gradients of our defense model, we replace $\text{sgn}(\cdot)$
and $\Pi_{\bm{x}'\in \Delta_{\bm{x}}}[\cdot]$ in~\eq{at_update} with their differentiable
approximations (see \secref{bpda_attack} for details).
We found that 
if the number of iterations (LL-PGD steps) for applying~\eq{at_update}
is low (i.e., $\le 2$), BPDA indeed enables the adversary to find valid adversarial examples. 
But when the number of iterations is set moderately high (i.e., $>4$),
BPDA is greatly thwarted; the adversary can hardly find any valid adversarial example (see \secref{bpda_attack}). 
This is because the differentiable approximations must be applied in each iteration,  
and as the number of iterations increases, the approximation error accumulates rapidly.

\paragraph{Finite difference gradients.}
Another strategy for estimating $\text{D}\tilde{g}(\bm{x})$ is the classic
finite difference estimation. Each element in the Jacobian matrix $\text{D}\tilde{g}(\bm{x})$
can be estimated using
\begin{equation}\label{eq:fde}
    \frac{\partial \tilde{g}_i(\bm{x})}{\partial \bm{x}_j}
    \approx \frac{1}{2h}\left[ \tilde{g}_i(\bm{x} + \bm{h}_j) - \tilde{g}_i(\bm{x} - \bm{h}_j)\right],
\end{equation}
where $\tilde{g}_i(\bm{x})$ indicates the $i$-th element (or pixel) of the transformed image, 
and $\bm{h}_j$ is a vector with all zeros except the $j$-th element (or pixel) which has a value $h$.

\begin{figure}[t]
  \centering
    \includegraphics[width=0.99\columnwidth]{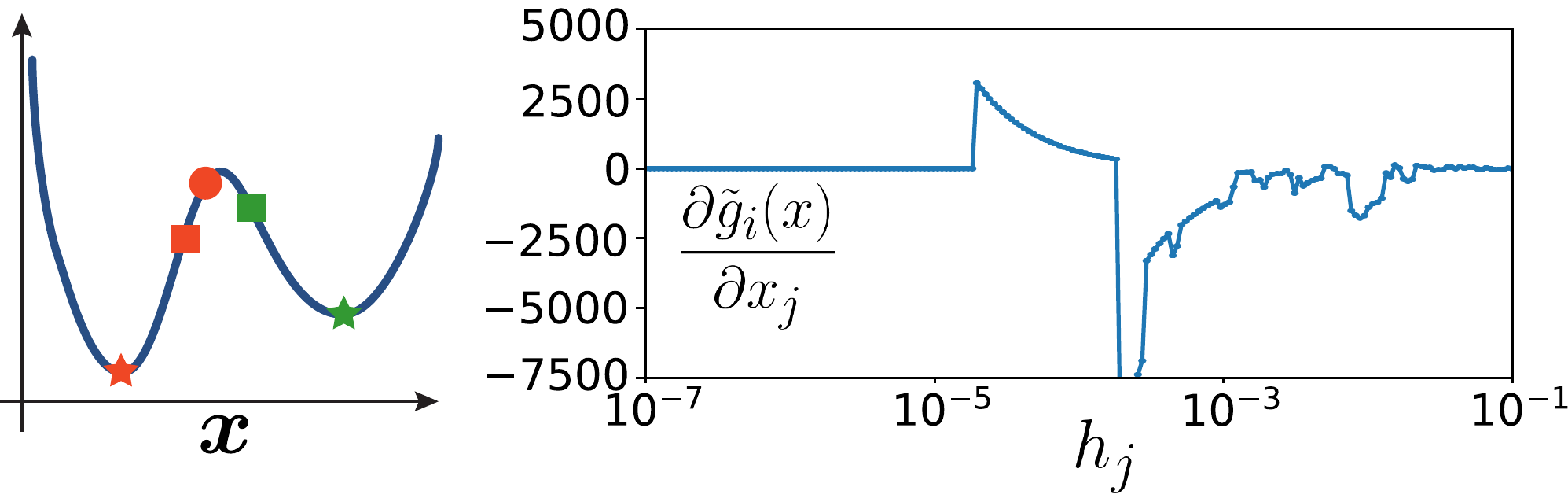}
  \vspace{-1mm}
  \caption{
      \textbf{(left)} 
      We show a 1D depiction of our input transformation $g(\bm{x})$, 
      the process aiming to find the local minimum of the optimization problem~\eq{attack_b}.
      Starting from an $\bm{x}_0$ at the red dot, 
      $g(\bm{x})$ will reach a position at the red star.
      Perturbing $\bm{x}_0$ toward one side (to the red square), $g(\bm{x})$ will still reach
      the red star, and in this way, the finite difference gradient vanishes.
      But if $\bm{x}_0$ is perturbed to the green square, 
$g(\bm{x})$ will reach the green star---an entirely different local minimum, 
and the finite difference gradient explodes.
\textbf{(right)}
  We plot $\frac{\partial \tilde{g}_i(\bm{x})}{\partial \bm{x}_j}$ (for particular $i$ and $j$ here) 
  estimated by finite difference method~\eq{fde} with an increasing $h_j$. 
  When $h_j$ is extremely small ($<10^{-5}$), the estimated gradient vanishes; as $h_j$ increases,
  the estimated gradient fluctuates severely, due to the reason illustrated on the left.
  }\label{fig:sensitivity}
  \vspace{-3mm}
\end{figure}

Our defense inherently thwarts this attacking strategy. 
It causes the adversary to suffer from either \emph{exploding} or \emph{vanishing} gradients~\cite{athalye2018obfuscated}.
Figure~\ref{fig:sensitivity}-left shows a 1D depiction illustrating this phenomenon in our method. 
Indeed, our experiments confirm that it is too unreliable to estimate
derivatives using~\eq{fde} (see \figref{sensitivity}-right).

\paragraph{Reparameterization.}
Vanishing and exploding gradients have been exploited as a defense
mechanism~\cite{song2018pixeldefend, samangouei2018defensegan}.
Yet those defenses have been proven vulnerable under a reparameterization strategy~\cite{athalye2018obfuscated}.
This strategy aims to find some differentiable function $h(\cdot)$ for 
a change-of-variable $\bm{x}=h(\bm{z})$ such that $\tilde{g}(h(\bm{z})) \approx h(\bm{z})$.
If such a function $h(\cdot)$ can be found, then 
one can compute the gradient of the differentiable function $f(h(\bm{z}))$ to launch adversarial attack.

To break our defense using this strategy, one must find an $h(\cdot)$ 
that constructs the adversarial examples of $f_b$ directly (so that $\tilde{g}(h(\cdot))=h(\cdot)$), without solving the optimization
problem~\eq{attack_b}. We argue that finding such an $h(\cdot)$ is extremely hard. 
If $h(\cdot)$ could be constructed, we would have a direct way of crafting adversarial examples;
PGD-type iterations would not be needed; and the entire territory of adversarial learning
would be redefined---which are unlikely to happen.

Indeed, we implemented this strategy by
training a neural network model $h_{\bm{\theta}}$
that aims to minimize $\|h_{\bm{\theta}}(\bm{x})-\tilde{g}(\bm{x})\|_2$
over the natural image distribution.
This attempt is futile.
Our experiments show that the generalization error of the trained $h_{\bm{\theta}}$
is too high to launch any valid adversarial attack (see \appref{repara}).
This conclusion also echos the prior studies~\cite{baluja2018learning,xiao2018generating},
which show that learning-based adversarial attacks usually
perform worse than gradient-based attacks. 

\paragraph{Identity mapping approximation.}
Some prior defense methods also use an optimization process to transform the input 
image---for example, the optimization that aims to erase
adversarial noise from the input image~\cite{guo2018countering,samangouei2018defensegan}.  
In those defenses, the transformed image $g(\bm{x})$ remain similar to the input $\bm{x}$.
Consequently, as shown in~\cite{athalye2018obfuscated}, those defenses can be
easily circumvented by replacing $g(\cdot)$ with the identity mapping in the backward pass
of BPDA attack.

Similarly, in our defense, if the perturbation range $\Delta$ in~\eq{at_update} 
for defining $g(\cdot)$ were set small, 
$g(\bm{x})$ (the adversarial example of $f_b$) would 
be close to $\bm{x}$, and our defense would be at risk. 
To prevent this vulnerability, we must ensure that $g(\cdot)$ be far from the identity mapping.
This requires us to set a relatively large $\Delta$.
In practice, we use $\Delta=0.2$ for pixel values ranging in $[0,1]$ 
(see details in \secref{bpda_attack}).

It turns out that a relatively large $\Delta$ is necessary but not sufficient.
The choice of the network model $f_b$ also affects how far $g(\bm{x})$ is from $\bm{x}$
statistically, 
as we will discuss next.

\begin{figure}[t]
  \centering
    \includegraphics[width=0.88\columnwidth]{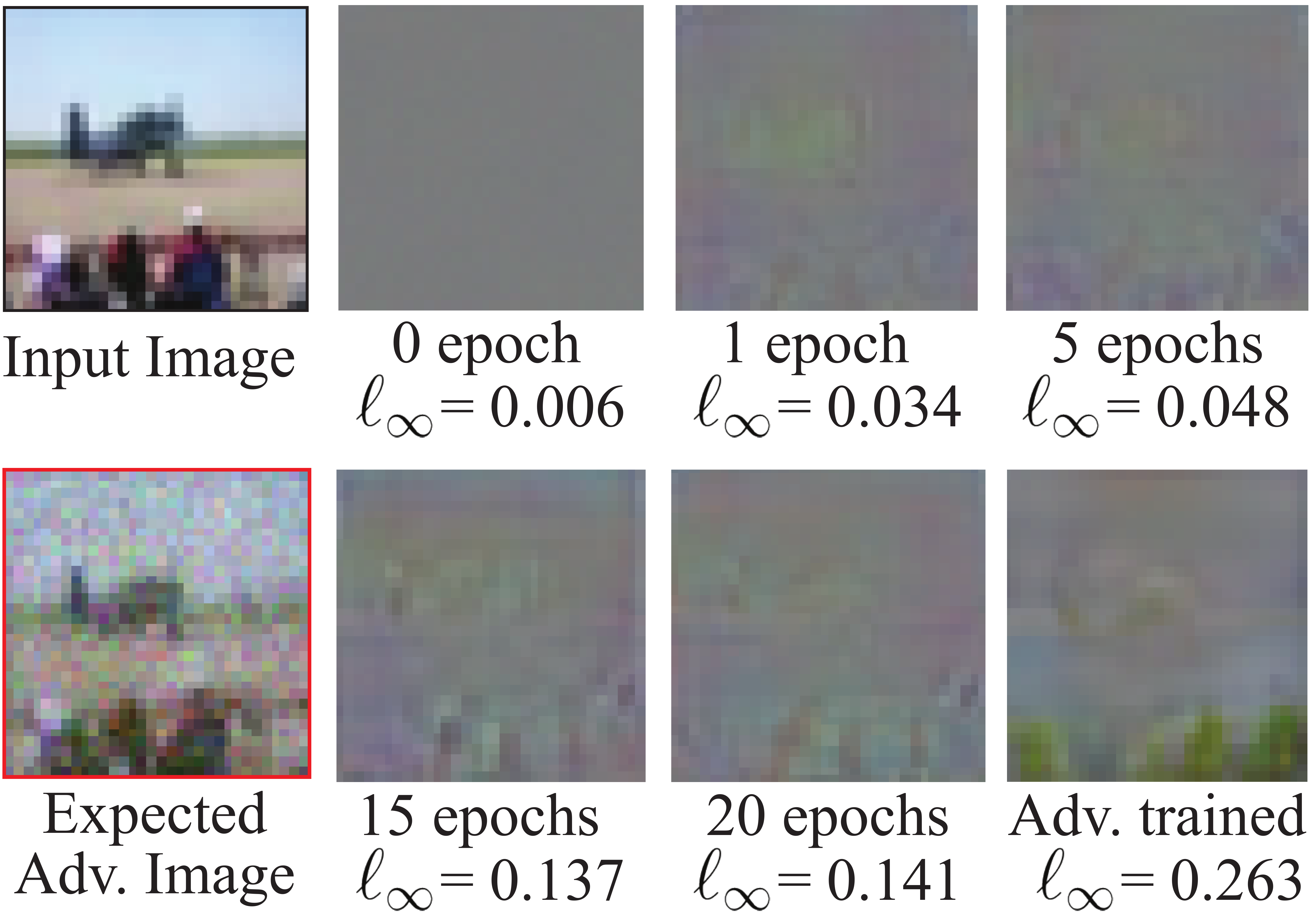}
    \vspace{-1mm}
    \caption{
We apply the adversarial transformation $g(\cdot)$ defined on different models $f_b$ 
to an input image (top-left). 
Corresponding to the six images toward the right 
are the $f_b$ models (with the same network structure) untrained, trained with an increasing number of epochs, and adversarially trained.
In each of those six images,
we visualize the normalized difference between the input $\bm{x}$ and 
the expectation over transformation (EOT) image
$\E_{\tilde{g}\sim\mathcal{T}}\tilde{g}(\bm{x})$ (estimated using 5000 samples).
The $L_\infty$ norms of the difference images are shown under the images.
Image in the red box (bottom-left) is the EOT image
$\E_{\tilde{g}\sim\mathcal{T}}\tilde{g}(\bm{x})$ produced using an adversarially trained $f_b$ model.
Because we intentionally use a large perturbation range $\Delta=0.2$, 
this image has pronounced artifacts.
    }\label{fig:expected_img}
    \vspace{-4mm}
\end{figure}

\subsection{Choosing Pre-trained Model $\bm{f_b}$}\label{sec:f_b}
\secspace

A large perturbation range $\Delta$ allows our adversarial transformation $g(\cdot)$
to output an image far from the input. Yet, because of the randomness in $g(\cdot)$, 
a large $\Delta_{\bm{x}}$ can not guarantee that $g(\bm{x})$ is \emph{statistically} different
from $\bm{x}$. 
If the expectation over the
transformation $\E_{\tilde{g}\sim\mathcal{T}}\tilde{g}(\bm{x})$ remains close
to $\bm{x}$, our defense method may still suffer from the aforementioned BPDA attack, in which
identity mapping can be used to approximate $g(\cdot)$ in the backward pass.

\begin{table*}[t]
\centering
\resizebox{0.91\textwidth}{!}{
\begin{tabular}{l|cccccccc}
\bottomrule
             & Untrained  & 1 epoch  & 2 epochs & 5 epochs & 10 epochs& 15 epochs & 20 epochs & Adv. trained \\ \hline
Standard Acc. &   81.8\%  &  81.6\%  & 82.4\%   & 83.0\%   &  82.4\% &  82.2\% & 82.7\%    &  82.9\%   \\
BPDA-I Acc.&      30.6\%  &  30.7\%  & 40.0\%   & 46.4\%   &  62.3\%  & 63.1\%& 62.9\%    &  80.5\%        \\
Avg. $L_{\infty}$ dist. &     0.005   &   0.033  &  0.036     & 0.067     &  0.130    & 0.133& 0.136      &  0.272      \\
\toprule
\end{tabular}}
\caption{ \textbf{Discovery for choosing $f_b$.}
Corresponding to individual columns
are $f_b$ models 
untrained, trained with an increasing number of epochs, and adversarially trained.
For the defense model $f_a(g(\cdot))$ equipped with each $f_b$, we evaluate its standard accuracy (first row) and 
robust accuracy (second row) under the BPDA-I attack (see \secref{bpda_attack}).
The third row shows $\|\bm{x}-\E_{\tilde{g}\sim\mathcal{T}}\tilde{g}(\bm{x})\|_\infty$ 
where $\E_{\tilde{g}\sim\mathcal{T}}\tilde{g}(\bm{x})$ is estimated using 5000 samples. 
Notice the correlation between the increase of the $L_\infty$ distance and the increase of adversarial robustness.
}\label{tab:expected_img}
\vspace{-3mm}
\end{table*}

This intuition is supported by an empirical discovery.
We experimented with an input transformation ${g}(\cdot)$ 
constructed using an \emph{untrained} model $\bar{f}_b$ whose weights are
assigned randomly. As shown in \figref{expected_img} and the first column in \tabref{expected_img}, 
the expectation over transformation $\E_{\tilde{g}\sim\mathcal{T}}\tilde{g}(\bm{x})$ is indeed close to $\bm{x}$ (in $L_\infty$ norm),
and the BPDA attack {with identity mapping approximation} can easily fool this defense model.

Next, we train a series of models $f^{(i)}_b$, each obtained with an increasing
number of training epochs. We found that as the number of training epochs increases, 
the expectation over transformation
$\E_{\tilde{g}\sim\mathcal{T}}\tilde{g}(\bm{x})$ resulted by using each of these $f^{(i)}_b$ models
drifts further away from $\bm{x}$, that is, 
$\|\bm{x}-\E_{\tilde{g}\sim\mathcal{T}}\tilde{g}(\bm{x})\|_\infty$ increases.
Meanwhile, the defense model $f(g(\cdot))$ trained with the corresponding $f^{(i)}_b$ 
becomes more robust, yielding
increasingly better robust accuracy under the BPDA attack (see \tabref{expected_img}).


Remarkably, we discover that an even larger distance
$\|\bm{x}-\E_{\tilde{g}\sim\mathcal{T}}\tilde{g}(\bm{x})\|_{\infty}$ can be obtained,
if the model $f_b$ is \emph{adversarially} trained.
When used in $g(\cdot)$, the adversarially trained model $f_b$
further improves the robustness of our defense model. 
This discovery confirms our intuition. 

\paragraph{Computational performance.}
The choice of $f_b$ also affects the computational cost of our defense method.
A complex network structure of $f_b$ makes $g(\cdot)$ expensive,
which in turn imposes a large performance overhead on both
the training of $f_a$ and the inference using $f_a$.
Therefore, a simple network structure is preferred.

The freedom of choosing a simple network $f_b$ brings our method a performance advantage
over adversarial training.
In adversarial training, adversarial examples are crafted on the classification network $f_a$
for each input image at every epoch. 
In our training, however, by choosing a model $f_b$ simpler than $f_a$,
it becomes faster to find adversarial examples. 
As shown in our experiments (in \secref{comp}), 
in comparison to adversarial training,
our defense 
requires shorter training time, and
at the same time offers stronger robustness.

\paragraph{Guiding rules.}
In summary, we present two guiding rules for choosing $f_b$.
\textbf{1)} $f_b$ should be chosen to yield a large $\|\bm{x}-\E_{\tilde{g}\sim\mathcal{T}}\tilde{g}(\bm{x})\|_{\infty}$ value.
Given an $f_b$'s network structure, adversarial training on $f_b$ (in pre-training step) is preferred. 
\textbf{2)} Meanwhile, the structure of $f_b$ should be as simple as possible.
In \appref{appfb}, we report $f_b$'s network structure that we use in our experiments.

\section{Devil's Advocate}\label{sec:eval}
\secspace

We now play devil's advocate in attacking our defense method.
In our defense, the network gradient with respect to the input (i.e., $\nabla f_a(g(\bm{x}))$)
is intentionally undefined. Thus one can not craft adversarial examples by directly applying
PGD-type methods on our defense (recall \secref{related}). We therefore evaluate our defense
against a range of other possible attacks, including those discussed in \secref{hardness}. 
Later in \secref{comp}, we will compare the \emph{worst-case} robustness of our defense 
under these attacks with various recently proposed defense methods. 


\paragraph{Common experiment setups.}
Experiments in this section are conducted on 
CIFAR-10 dataset~\cite{krizhevsky2009learning}
with standard training/test split.
We use ResNet18~\cite{he2016deep} as the classification model $f_a$
and a small VGG-style network for $f_b$, whose details are given in \appref{appfb}.
{All models are trained for 80 epochs using Stochastic Gradient Descent (SGD) (constant learning rate=0.1, momentum=0.9).}
Our adversarial transformation $g(\cdot)$ performs LL-PGD update~\eq{at_update} 
for 13 iterations, each with a stepsize $\epsilon=\Delta/6$.
The perturbation range $\Delta$ varies in individual experiments, and will be reported therein.

\paragraph{Metric.}
Following prior work, 
our evaluation uses an accuracy measure defined as the ratio of the number of correctly classified
images to the total number of tested images. 
We refer to this measure as \emph{standard accuracy}
if the tested images include only clean images, 
and as \emph{robust accuracy}
if the tested images consist of adversarially crafted images. 

\subsection{BPDA Attack and the Variants}\label{sec:bpda_attack}
\secspace
BPDA attack~\cite{athalye2018obfuscated}, as reviewed in \secref{hardness}, is a powerful  
way to estimate network gradients that are obfuscated by defense methods. The
estimated gradients are then used in PGD-type methods (if the defense is
deterministic) or the EOT method (if the defense is randomized) for crafting
adversarial examples. 
BPDA has circumvented a handful of recent defense techniques~\cite{guo2018countering,xie2018mitigating,ma2018characterizing,song2018pixeldefend,buckman2018thermometer}
that implement gradient obfuscation, in many defenses resulting in
0\% robust accuracy. We therefore evaluate our defense against it and its possible variants.


\begin{figure}[t]
  \centering
  \vspace{-2mm}
    \includegraphics[width=0.82\columnwidth]{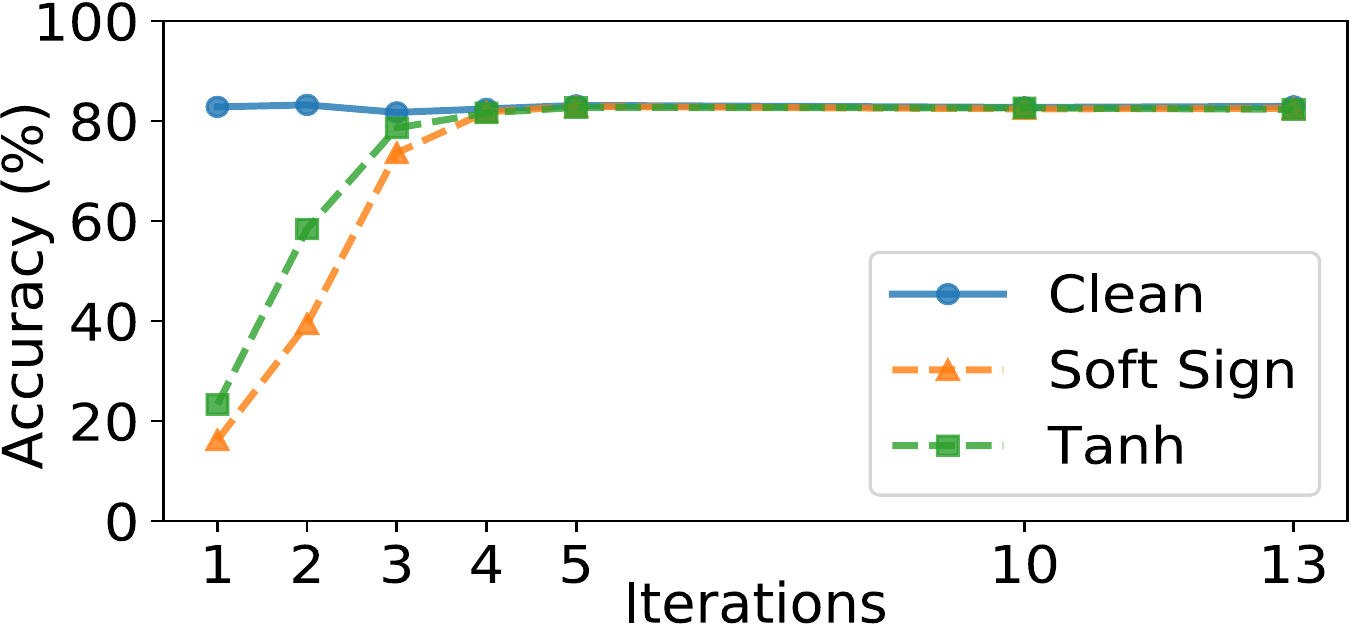}
    \caption{\textbf{Robustness under BPDA.}
  We evaluate the robust accuracy of our defense under two versions of BPDA attacks,
  which replace $\text{sgn}(\cdot)$ in~\eq{at_update} 
  with \textsf{soft sign} (in orange) and \textsf{tanh} (in green), respectively.
  The resulting two robust accuracies are compared with our defense model's standard accuracy (in blue) evaluated with 
  \textsf{clean} (natural) images.
  Along X-axis,
  we repeat this evaluation, each time with an increasing number of LL-PGD steps~\eq{at_update}
  in our adversarial transformation $g(\cdot)$.
  To highlight only the effect of the smooth approximations of $\text{sgn}(\cdot)$
  and $\Pi_{\bm{x}'\in \Delta_{\bm{x}}}(\cdot)$, 
  we factor out the randomness in our defense by disabling the random start 
  at the beginning of~\eq{at_update}.
  After the network gradient is estimated using BPDA,
  we use PGD to search 
  for adversarial examples with a maximum perturbation size of 0.031
  (in $L_\infty$ norm). The PGD search takes 50 iterations with a stepsize 0.002.
  \label{fig:std-bpda}}
  \vspace{-3mm}
\end{figure}

\paragraph{Differentiable approximation on backward pass.}
The update rule~\eq{at_update} in our adversarial transformation involves two
non-differentiable operators, namely, $\textrm{sgn}(\cdot)$ and $\Pi_{\bm{x}'\in \Delta_{\bm{x}}}(\cdot)$,
{whose specific forms are given in \appref{bpda}.}
To launch BPDA attack, we need to replace them with differentiable operators and 
compute their derivatives. We experimented with two different 
smooth approximations of $\textrm{sgn}(\cdot)$: the \emph{soft sign} function
$\frac{x}{1+|x|}$ and \emph{tanh} function  $\frac{e^x-e^{-x}}{e^x+e^{-x}}$.
The smooth approximation of $\Pi_{\bm{x}'\in \Delta_{\bm{x}}}(\cdot)$ is not explicitly 
defined. Instead, we directly approximate its derivative using
\begin{equation}\label{eq:proj_approx}
    \frac{\text{d}}{\text{d}x}\Pi_{\bm{x}'\in \Delta_{\bm{x}}}(x)\approx \begin{cases} 1, & \text{if}\ |x|<\Delta,\\
                                                \frac{1}{(1+|x|)^2}, & \text{otherwise.}
    \end{cases}
\end{equation}
Reported in \figref{std-bpda}, the experiments show that our defense is robust to this attack, 
as long as the number of LL-PGD steps $N$ in $g(\cdot)$ is not too small (i.e., $N>5$).

If the number of LL-PGD steps $N$ is set too small, BPDA attack is indeed
able to find adversarial examples (\figref{std-bpda}).
Therefore, another attempt one may ponder is to craft adversarial examples on 
a model trained with a small $N$ ($N\le3$), and use them to transfer attack 
our defense (which is trained with a larger $N$). This attack remains ineffective (see details in
\appref{bpda}). We conjecture that this is because the 
adversarial examples for the model with a small $N$
have a different distribution from that with a larger $N$~\cite{papernot2017practical,su2018is}.


\paragraph{Identity mapping approximation.}
Another possible attack is by replacing the input transformation $g(\cdot)$
with the identity mapping for gradient estimation in BPDA backward pass (recall
discussion in \secref{hardness}). We refer this attack as BPDA-I attack.
Under this attack, several previous defenses (e.g.,~\cite{samangouei2018defensegan,song2018pixeldefend,guo2018countering})
have been nullified.

\begin{figure}[t]
  \centering
  \includegraphics[width=0.80\columnwidth]{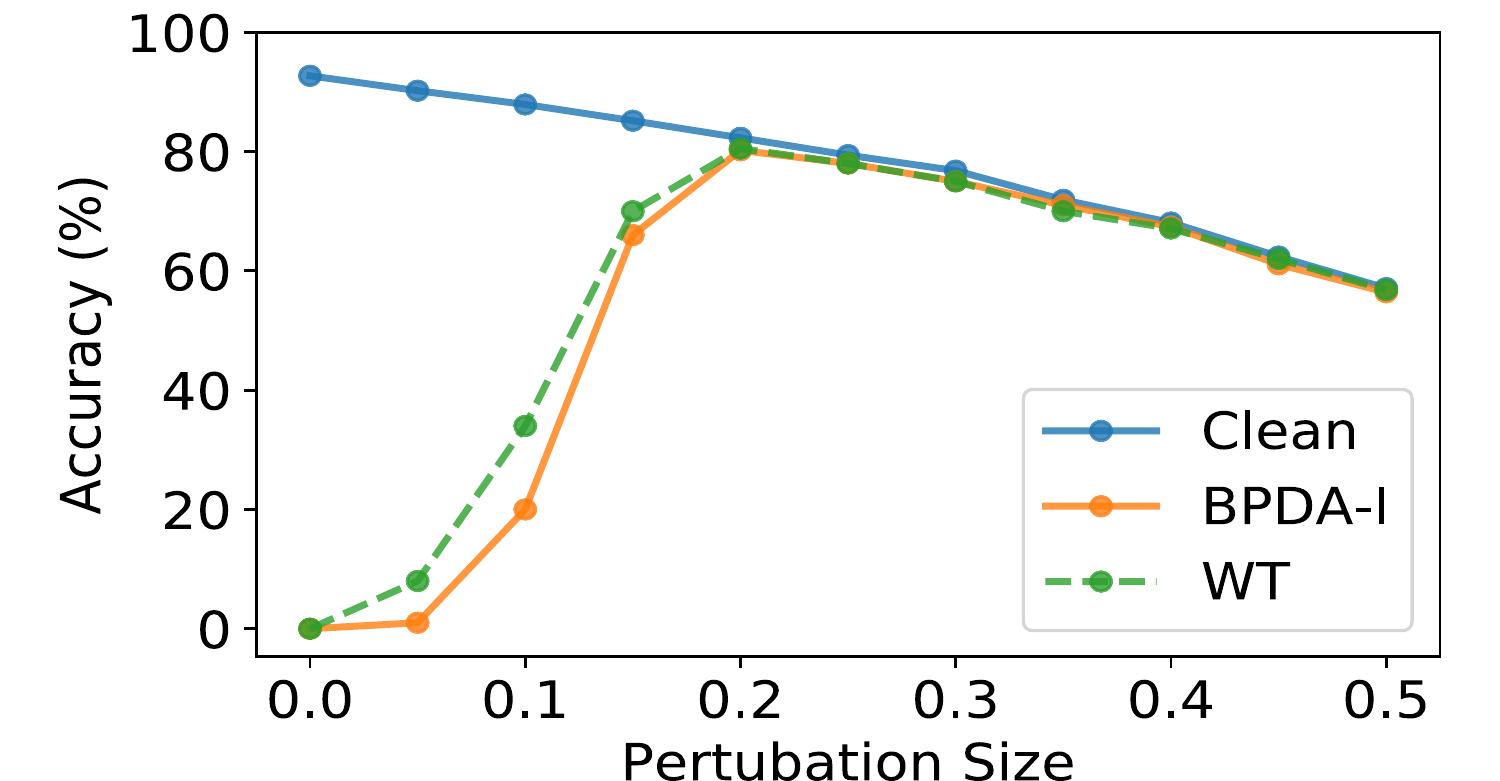}
  \caption{\textbf{Robustness w.r.t.~perturbation size.}
  We test our defense robustness using independently trained defense models with increasing 
  perturbation range $\Delta$ used in $g(\cdot)$.
  The standard accuracy (blue curve) is measured using \textsf{clean} images.
  The robust accuracies under \textsf{BPDA-I} attack (\secref{bpda_attack}) 
  and \textsf{WT} attack (\secref{transfer}) are in orange and green, respectively.
  In our input transformation, we enable random start of~\eq{at_update}.
  Therefore, in BPDA-I attack, we use 500 samples of $\tilde{g}(\cdot)$ for EOT gradient estimation.
  The gradient-descent setup is the same as that in the earlier experiments in \figref{std-bpda}.
  \label{fig:std-bpda-da}}
  \vspace{-5mm}
\end{figure}

We applied BPDA-I attack on our defense. The attack setup and results are summarized 
in \figref{std-bpda-da}. 
As the perturbation size $\Delta$ in the adversarial transformation
increases, the robust accuracy of our method increases. The robust accuracy is always upper bounded
by the standard accuracy, which decreases gradually as $\Delta$ increases.
If $\Delta$ is too large, the excessive perturbations to the input make 
the network $f_a$ harder to learn and thus lower the standard and robust accuracies.
Empirically, $\Delta=0.2$ offers the best performance.

\paragraph{Reparameterization.}
In \secref{hardness}, we described another
BPDA strategy, one that uses reparameterization to smoothly approximate 
our adversarial transformation $g(\cdot)$. As discussed therein, it is extremely hard 
to directly derive the reparameterization function. 
Instead, we attempted to train a Fully Convolutional Network~\cite{long2015fully}
to represent $h(\bm{z})$. We denote this network as $h(\bm{x};\bm{\theta})$, whose weights 
are optimized with the loss function,
$\ell(\bm{\theta})=\E_{\bm{x}\in\mathcal{X}}\norm{h(\bm{x};\bm{\theta})-g(\bm{x})}^2.$
Here $\mathcal{X}$ represents the distribution of natural images (we use CIFAR-10 
as the training dataset).

Our experiment shows that although we can reach a low loss value in training $h(\bm{x};\bm{\theta})$,
the loss on test dataset always stays high, indicating that $h(\bm{x};\bm{\theta})$
is always overfitted. As a result, the adversarial examples resulted in this way have
almost no effect on our defense model---the accuracy drop under this attack
is within 1\% from the standard accuracy. 
See \appref{repara} for the details of this experiment.

\subsection{Gradient-Free Attacks}\label{sec:transfer}
\secspace
Several attacking methods require no gradient information of the
model, and they can be employed to potentially threaten our defense.
As discussed in \secref{related},
these attacks fall into two categories: transfer attack and query-based attack.
Against both types of attacks we evaluate our defense.


%

\paragraph{White-box transfer attack.}
In white-box setting, the adversary has full knowledge
of our defense model. A tempting idea is to generate adversarial examples on
the classifier model $f_a$, and use them to transfer attack our defense
model $f_a(g(\cdot))$.
Note that this differs from BPDA-I attack in \secref{bpda_attack},
where $f_a$ is used only in the backward pass for gradient estimation
while the forward pass still uses the full model $f_a(g(\cdot))$.
Here, in contrast, adversarial examples are generated solely on $f_a$. 
We refer this attack as White-box Transfer (WT) attack, 
and report the robust accuracies of our defense in \figref{std-bpda-da},
along with the results under BPDA-I attack. We found that our model's robustness performances
under both attacks are similar, and $\Delta=0.2$ is the best choice. 

One may realize another attacking possibility by noticing the way we choose $f_b$ (on which we perform adversarial transformation).  
In \secref{f_b}, we present that $f_b$ should be chosen such that for a given natural image $\bm{x}$
the average transformation $\E_{\tilde{g}\sim\mathcal{T}}\tilde{g}(\bm{x})$ stays far from $\bm{x}$.
Thus, it seems plausible to first generate adversarial examples on $f_a$ using PGD attack starting from 
the average transformation $\E_{\tilde{g}\sim\mathcal{T}}\tilde{g}(\bm{x})$, and use them to attack our full model $f_a(g(\cdot))$.
However, thanks to the large perturbation range $\Delta_{\bm{x}}$ we use (recall \secref{hardness}),
$\E_{\tilde{g}\sim\mathcal{T}}\tilde{g}(\bm{x})$ is always far from a natural
image (see the image in the red box of \figref{expected_img}).
Thus the adversarial examples generated in this way 
all have easily noticeable artifacts; they are not valid.

\begin{table}[t]
\centering
\resizebox{1.0\columnwidth}{!}{ %
\begin{tabular}{l|cccc}
\bottomrule
Defense Model                       & Clean  & Transfer  & HSJ & GA \\ \hline
No defense                          & 92.9\% & 1.3\%    &  3.6\%     & 5.8\%          \\
Madry et al.~\cite{madry2017towards}                        & 81.7\% & 77.5\%    &  72.1\%     & 78.4\%          \\
Ours                                & 82.9\% & \textbf{78.1\%}    &  \textbf{82.0\%}     & \textbf{81.9\%}          \\
\toprule
\end{tabular}}
\caption{\textbf{Robustness under black-box attacks.}
In the transfer attack, adversarial examples are crafted on an independently trained ResNet18 model.
The query-based attacks are performed using a third-party library \textit{foolbox}~\cite{rauber2017foolbox} with 
default parameters to launch these attacks.
All adversarial examples are restricted in the $L_{\infty}$ ball with a perturbation size of $0.031$.
}\label{tab:black}
\vspace{-4mm}
\end{table}

\paragraph{Black-box attacks.}
We also evaluate our defense against the black-box attacks, 
including the black-box transfer attack~\cite{papernot2017practical}
and two most recently introduced query-based attacks, HopSkipJumpAttack (HSJ)~\cite{chen2019boundary} and GenAttack (GA)~\cite{alzantot2019genattack}.
In \tabref{black}, we summarize
the robust accuracies of our model under these attacks, 
along with two baselines from the same classification model ($f_a$)
optimized respectively using standard training and adversarial training.

\secspace
\section{Comparisons and Further Evaluation}\label{sec:comp}
\secspace

\paragraph{Comparisons.} 
We now compare the robustness of our method with other state-of-the-art defense
methods under white-box attacks. Unlike many others that can be attacked using PGD-type methods, 
our defense model is inherently non-differentiable, immune to direct PGD attacks.
Therefore it is not possible to compare all these defense methods under exactly the same attacks. Instead, we 
compare the \emph{worst-case} robustness of our method under all the attacks described in \secref{eval} with other methods.

\begin{table}[t]
\centering
\begin{tabular}{l|cccc}
\bottomrule
Method & $A_{std}$ & $A_{rob}$ & Best Attack \\ \hline
No defense &  92.9\% & 0.0\% &PGD \\
Madry et al.~\cite{madry2017towards} & 81.7\% & 42.7\% & PGD      \\
Zhang et al.~\cite{zhang2019theoretically} & 80.4\% & 44.6\% &PGD \\
Xie et al.~\cite{xie2019feature}& 83.8\% & 45.2\% & PGD           \\
Guo et al.*~\cite{guo2018countering}&  - & 0.0\%  & BPDA          \\
Buckman et al.*~\cite{buckman2018thermometer}&-& 0.0\%& BPDA        \\
Dhillon et al.*~\cite{s2018stochastic}&-& 0.0\% & BPDA \\
Song et al.*~\cite{song2018pixeldefend}&-& 5.0\% & BPDA \\
Ours (under BPDA)& 82.9\% & \textbf{80.2\%} & BPDA \\
Ours& 82.9\% & \textbf{78.1\%} & Transfer \\
\toprule
\end{tabular}
\caption{\textbf{Comparisons on CIFAR-10.} 
Methods indicated by * are those circumvented in~\cite{athalye2018obfuscated}.
We evaluate other methods using the code provided in the original papers, 
training them using the same network and hyperparameters as our method.
The perturbation range of all adversarial examples is $\Delta=0.031$.  
The last column indicates the most efficient attacking method that produces the worst robustness.
The second last row indicates the worst-case robustness of our method under all BPDA-type attacks,
while the last row indicates our worst-case robustness under all attacks.
}\label{tab:cmp_cifar}
\vspace{-3mm}
\end{table}



The comparison results on CIFAR-10 dataset are summarized in \tabref{cmp_cifar},
where $A_{std}$ is the standard accuracy tested with clean images,
and $A_{rob}$ is the \emph{worst-case} robust accuracy 
under all tested attacks. The methods indicated by a star (*) are those circumvented
by Athalye et al.~\cite{athalye2018obfuscated}. We include their results therein as a reference.
The other defense methods (including ours) all use ResNet18 as their classification model, trained
with SGD (learning rate=0.1, momentum=0.9) for 80 epochs.

On CIFAR-10 dataset, the most effective attack on our method is the {black-box} transfer attack
(\secref{transfer}), although its severity surpasses BPDA attacks only
slightly: the worst-case robust accuracy of our method 
under BPDA attack and its variants (\secref{bpda_attack}) is 80.2\%.
Nevertheless, our robustness performance is significantly better than
the state-of-the-art methods, as shown in \tabref{cmp_cifar}.

We also performed the comparisons on Tiny ImageNet dataset, and our method
demonstrates significantly stronger robustness as well.
In short, our worst-case robust accuracy is \textbf{40.2\%}, in stark contrast to previous 
methods, which all have robust accuracies around \textbf{18\%}.
The results are reported in details in \appref{tiny}.


\paragraph{Training cost.}
Our method has significantly lower training cost than 
the adversarial training~\cite{madry2017towards,zhang2019theoretically,xie2019feature}, while offering stronger robustness.
For example, our method takes \textbf{82 minutes}
to train a ResNet18 model on CIFAR-10 for 80 epochs, 
while the adversarial training takes \textbf{460 minutes}. 
As a baseline, the standard training takes 56 minutes.  
All timings are measured on a NVIDIA RTX 2080Ti GPU.

\begin{figure}[t]
  \centering
    \includegraphics[width=0.86\columnwidth]{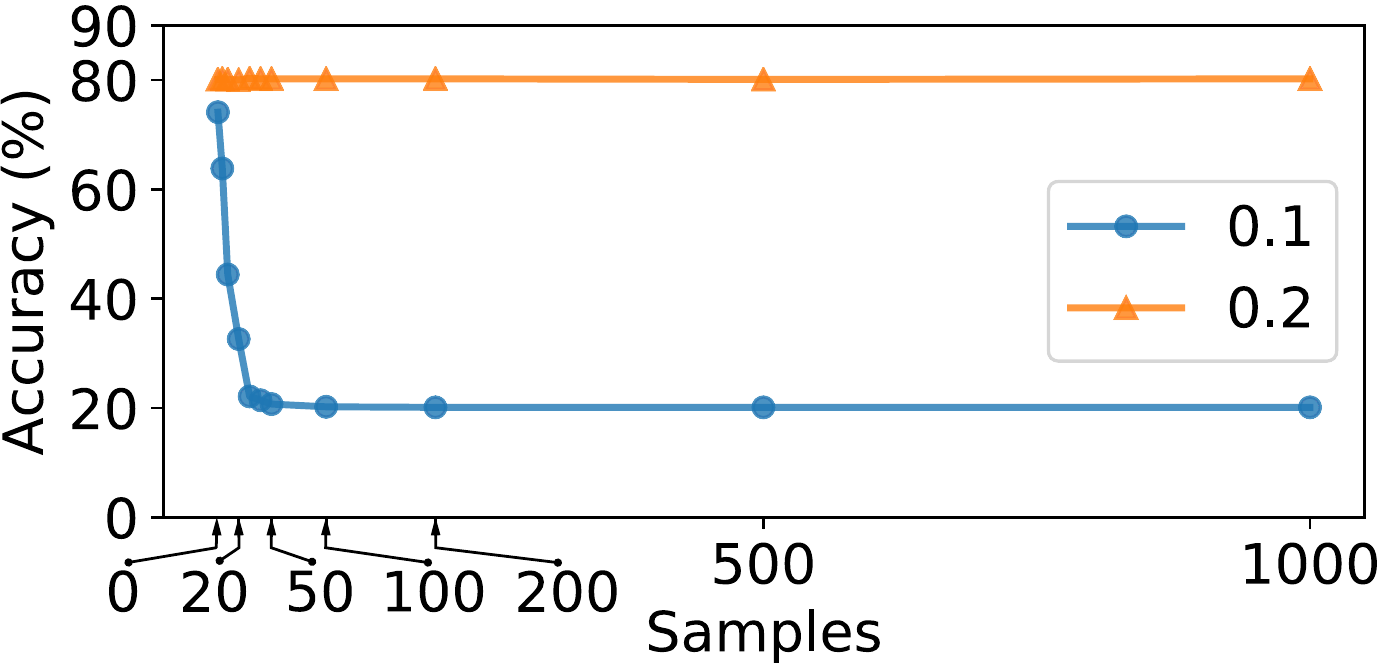}
    \vspace{-1mm}
\caption{\textbf{Robustness w.r.t.~EOT samples.}
When using EOT to attack our method, we sample our adversarial transformation with
different random starts $\bm{x}+\bm{\delta}$ to estimate the expected $\nabla f_a(g(\bm{x}))$.
When $\Delta=0.1$ (blue curve), increasing the sample size allows EOT to better attack our defense, 
until it plateaus. But when $\Delta=0.2$ (orange curve), EOT becomes persistently inefficient.
}\label{fig:eot}
    \vspace{-4mm}
\end{figure}

\paragraph{Sufficiency of EOT samples.}
Our defense is randomized, and when using EOT to attack our method we 
take 500 samples of $\tilde{g}(\cdot)$ (recall experiments in \figref{std-bpda-da}).
Here we conduct additional experiments to ensure the sufficiency of using 500 samples for estimating the expectation.
As shown in \figref{eot}, when the perturbation size $\Delta$ is small and 
the number of samples is also small (e.g., $<100$), increasing sample size indeed allows EOT 
to better attack our method. However, when the perturbation size is set to 0.2, the value
we consistently use throughout all our evaluations,
EOT attacks became persistently inefficient, regardless of the sample size.
Therefore, we conclude that 500 samples in EOT allow thorough evaluation of our defense against EOT.

\secspace
\section{Conclusion}
\secspace
We have presented a simple defense mechanism against adversarial attacks.
Our method takes advantage of the numerical recipe 
that searches for adversarial examples, and turns this harmful process into a useful input 
transformation for better robustness of a network model. 
On CIFAR-10 and Tiny ImageNet datasets, it demonstrates
state-of-the-art worst-case robustness under a wide range of attacks. 
We hope our work can offer other researchers a new perspective to study the adversarial defense mechanisms.
In the future, we would like to better understand the theoretical properties of our
adversarial transformation and their connections to stronger adversarial robustness.

{\small
\bibliographystyle{ieee_fullname}
\bibliography{ref}
}

\clearpage
\setcounter{page}{1}

\begin{strip}
    \vspace{-6mm}
\begin{center}
 \Large
 \textbf{Supplementary Document}\\ 
 \smallskip
 \smallskip
 \textbf{One Man's Trash is Another Man's Treasure:\\ Resisting Adversarial Examples by Adversarial Examples}
 \smallskip
\end{center}
\end{strip}

\appendix
\section{Additional Experiments and Setups}
\subsection{Network Structure of $\bm{f_b}$}\label{sec:appfb}
Following the guidelines presented at the end of \secref{f_b},
we choose to use a VGG-style small network in $f_b$ for defining our adversarial transformation.
This network is simple enough to enable fast adversarial transformation,
while producing the expectation over transformation image 
(i.e., $\E_{\tilde{g}\sim\mathcal{T}}\tilde{g}(\bm{x})$) drastically different from the input image. 
The structure of $f_b$ for experiments on CIFAR-10 dataset is described in \tabref{arch_fb}.

On Tiny ImageNet dataset, the network structure of $f_b$ remains largely the same
except two minor changes to accommodate the different resolution of the images in Tiny ImageNet.
Namely, the changes are at the 12nd layer (which has a dimension 2048) and the 15th layer (which has a dimension 200).

\begin{table}[h]
\centering
\begin{tabular}{l|ll}
\bottomrule
Layer & Module & Output Size \\ \hline
1 & Input & 3$\times$32$\times$32 \\
2 & Conv(k=3), BN, ReLU & 64$\times$32$\times$32 \\
3 & MaxPool & 64$\times$16$\times$16\\
4 & Conv(k=3), BN, ReLU & 128$\times$16$\times$16 \\
5 & MaxPool & 128$\times$8$\times$8 \\
6 & Conv(k=3), BN, ReLU & 128$\times$8$\times$8 \\
7 & Conv(k=3), BN, ReLU & 128$\times$8$\times$8 \\
8 & MaxPool & 128$\times$4$\times$4 \\
9 & Conv(k=3), BN, ReLU & 128$\times$4$\times$4 \\
10 & Conv(k=3), BN, ReLU & 128$\times$4$\times$4 \\
11 & MaxPool & 128$\times$2$\times$2 \\
12 & Flatten & 512 \\
13 & Linear, ReLU, Dropout& 512 \\
14 & Linear, ReLU, Dropout& 512 \\
15 & Linear (output)& 10 \\

\toprule
\end{tabular}          
\caption{\textbf{Network structure for $f_b$}. Here BN denotes batchnorm operation,
and Conv(k=3) denotes convolutional layer with a kernel size of 3.}\label{tab:arch_fb}
\end{table}

\subsection{Reparameterization Attack}\label{sec:repara}
As discussed in \secref{hardness} and \ref{sec:bpda_attack},
to launch the reparameterization attack,
we need to find a forward function $h(\cdot)$ that approximate our adversarial transformation process.
To this end, we attempted to train a Fully Convlutional Network (FCN)~\cite{long2015fully},
denoted as $h(\bm{x};\bm{\theta})$, through the following optimization,
\begin{equation}\label{eq:re}
    \bm{\theta} = \argmin_{\bm{\theta}} \mathbb{E}_{\bm{x} \in \mathcal{X},\bm{\delta}\in\bm{\Delta}} \norm{h(\bm{x}+\bm{\delta};\bm{\theta})-\tilde{g}_{\bm{\delta}}(\bm{x})}^{2},
\end{equation}
where $\mathcal{X}$ is the given dataset, $\bm{\delta}$ is the initial input perturbation 
in the $L_\infty$ ball of size $\Delta$ (as described in \secref{hardness}),
$\tilde{g}_{\bm{\delta}}(\cdot)$ is the deterministic version of our adversarial transformation $g(\cdot)$:
it starts the adversarial search iteration~\eq{at_update} from $\bm{x}+\bm{\delta}$ by using a sampled $\bm{\delta}$.

\begin{figure}[t]
  \centering
  \vspace{-2mm}
    \includegraphics[width=0.82\columnwidth]{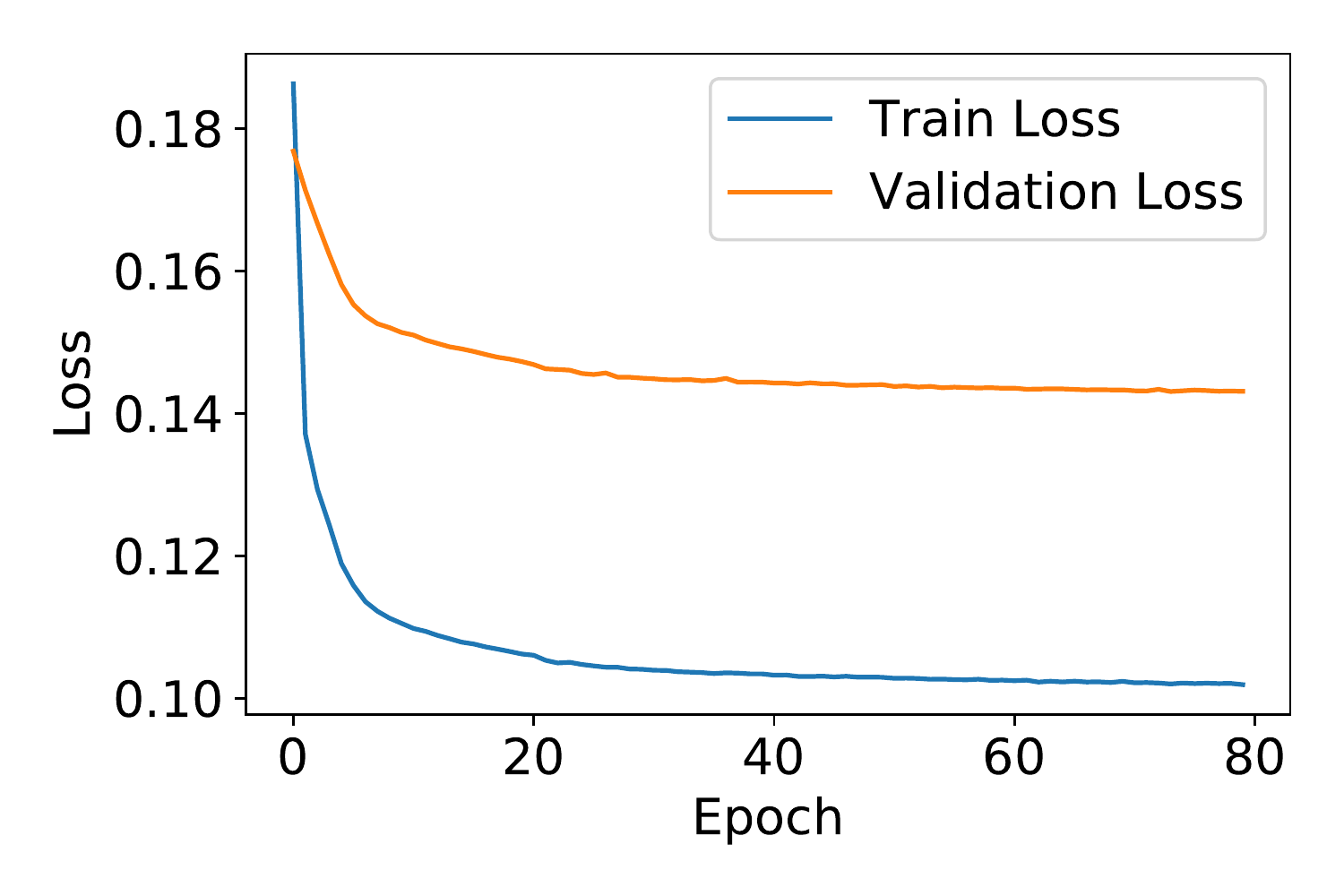}
    \vspace{-1mm}
    \caption{\textbf{Training Loss and validation loss} in reparameterization attack.}\label{fig:re}
  \vspace{-3mm}
\end{figure}

However, after optimizing \eq{re}, we found that although the FCN model can 
reach a relatively low training error,
the error on test set remains high, as depicted in \figref{re}.
This suggests that the FCN model is not able to learn a $h(\bm{x};\bm{\theta})$ that generalizes well.
The inability to generalize is not a surprise: if $h(\bm{x};\bm{\theta})$ could generalize well, we would have a direct
way of crafting adversarial examples; and PGD-type iterations would not be needed---which are all unlikely.

Indeed, when we use the trained $h(\bm{x},\bm{\theta})$
to launch a reparameterization attack to our model, the attack hardly succeeds.
Under this attack (on CIFAR-10),
the robust accuracy of our defense is 81.1\%, even better than
the robust accuracy under BPDA-I attack (80.2\%).
In fact, this accuracy nearly reaches its upper bound, the standard accuracy (i.e., 82.9\%), as reported in \tabref{cmp_cifar}
of the main text.

\begin{table*}[t]
\centering
\begin{tabular}{c|ccccc}
\bottomrule
& &\multicolumn{3}{c}{Robust Acc. (transfer attack)} \\
Defense Model Iterations & Standard Acc.   & $N=1$ & $N=2$ & $N=3$ \\ \hline
$N=3$ &  81.7\% & 69.7\%   & 66.4\% & 79.5\% \\
$N=4$ &  82.4\% &  75.6\%   & 78.7\% & 79.9\%  \\
$N=5$ &  83.1\% &  81.6\% & 80.7\% & 81.7\%  \\
$N=10$ & 82.7\%  & 81.4\% & 81.2\% & 81.0\%  \\
$N=13$ & 82.9\%  & 80.9\%  &  81.1\% & 81.3\%  \\
\toprule
\end{tabular}          
\caption{\textbf{Transfer attack based on BPDA.} Each row shows the standard and robust
accuracies of our defense model with a different number of LL-PGD steps in its adversarial 
transformation $g(\cdot)$. The number of LL-PGD steps in our defense model is shown in the left most
column. The right most three columns correspond to the models with smaller numbers of LL-PGD steps.
We use these models to craft adversarial examples to transfer attack our defense model.
It shows that even when the number $N$ of LL-PGD steps in our defense model is moderately large ($N\ge5$),
the transfer attacks become ineffective.
In all the evaluations, the perturbation size $\Delta$ in our defense model is set as $\Delta=0.2$.
}\label{tab:bpda_num}
\end{table*}

\subsection{BPDA Attack Details and Additional Results}\label{sec:bpda}
As described in \secref{bpda_attack},
we evaluate our defense model under the BPDA attack.
To launch BPDA attack, we need to replace the non-differentiable
operators in our adversarial transformation with their smooth approximations.
In particular, the non-differentiable operators in the adversarial update
rule~\eq{at_update} (in the main text) are the $\textrm{sgn}(\cdot)$ function,
\begin{equation}\label{eq:sgn}
 \text{sgn}(\bm{x}) = \begin{cases} 1 & \text{if}\ \bm{x} > 0, \\
                  0, & \text{if}\ \bm{x}= 0,\\
                  -1 & \text{if}\ \bm{x}<0. 
    \end{cases}
\end{equation}
and the $L_\infty$ projection operator,
\begin{equation}\label{eq:clamp}
      \Pi_{\bm{x}'\in \Delta_{\bm{x}}}(\bm{x})=\begin{cases}
  \bm{x} & \text{if}\ |\bm{x}|\leq\Delta,\\
  -\Delta & \text{if}\ \bm{x}<-\Delta, \\
  \Delta & \text{if}\ \bm{x}>\Delta.
  \end{cases}
\end{equation}
In \secref{bpda_attack},
we experimented with two different smooth approximations of 
the $\textrm{sgn}(\cdot)$ function, namely, the soft sign function
$\frac{x}{1+|x|}$ and {tanh} function  $\frac{e^x-e^{-x}}{e^x+e^{-x}}$,
and the projection operator is replaced by directly approximating its
derivative using~\eq{proj_approx}.


Also discussed in \secref{bpda_attack} is an additional transfer attack:
First, we craft adversarial examples by setting the number $N$ of LL-PGD steps to be 
a small value. This is motivated by the observation that, as shown in \figref{std-bpda}, BPDA attack is able to 
find effective adversarial examples when $N$ is small.
We then use the resulting adversarial examples to transfer attack our defense model, which 
uses a larger number of LL-PGD steps in the adversarial transformation (in both training and inference). 
As summarized in \tabref{bpda_num},
our experiment shows that this attack remains ineffective to our defense.


\begin{table}[t]
\centering

\begin{tabular}{l|cccc}
\bottomrule
Method & $A_{std}$ & $R_{rob}$ & Best Attack \\ \hline
No defense &  58.2\% & 0.0\% &PGD \\
Madry et al.~\cite{madry2017towards} & 42.7\% & 17.3\% & PGD      \\
Zhang et al.~\cite{zhang2019theoretically} & 40.6\% & 17.7\% &PGD \\
Mao et al.~\cite{mao2019metric} &  40.9\%  &  17.5\% & PGD \\
Ours (Under BPDA)& \textbf{48.8\%} & \textbf{47.9\%} &  BPDA  \\
Ours& \textbf{48.8\%} & \textbf{40.2\%} &  Transfer  \\

\toprule
\end{tabular}          

\caption{\textbf{Comparisons on Tiny ImageNet.} 
The layout of this table is similar to \tabref{cmp_cifar} in the main text (i.e., the comparisons on CIFAR-10).
The perturbation range of all adversarial examples is $\Delta=0.031$.
The last column indicates the most efficient attacking method that produces the 
worst robustness. The second last row indicates the worst-case
robustness of our method under all BPDA-type attacks, while the
last row indicates our worst-case robustness under all attacks.
}\label{tab:cmp_tiny}
\vspace{-2mm}
\end{table}

\subsection{Evaluation on Tiny ImageNet}\label{sec:tiny}
We also evaluate our defense model on Tiny ImageNet dataset consisting of 
 64px$\times$64px RGB images. These images fall into 200 classes, each
has 500 images for training and 50 images for testing.
%
%
Following the evaluations setups in prior works, the adversarial examples used in the attacks 
have a maximum perturbation size (in $L_\infty$ norm) of 0.031 for pixel values ranging in [0,1].
We use ResNet18 as our classification network (in $f_a$) and the network structure of $f_b$ 
is described in \appref{appfb}.

We compare our method with the state-of-the-art method~\cite{mao2019metric} evaluated on Tiny
ImageNet and the methods based on adversarial training~\cite{madry2017towards,zhang2019theoretically}.  
For all those methods, we use the implementation code provided in their original papers.
When comparing with these methods, we use the same training protocol: 
the models are optimized use SGD (learning rate=0.1, momentum=0.9) and trained for 80 epochs.

As shown in \tabref{cmp_tiny},
our method demonstrates significantly stronger
robustness in comparison to previous methods. Our worst-case robust accuracy
is \textbf{40.2\%}. In contrast, previous methods 
have robust accuracies around \textbf{18\%}. 
Remarkably,
the standard accuracy of our method also outperforms previous methods.

\subsection{Expectation over Transformation Images}\label{sec:appexpected}
\begin{figure*}[t]
  \centering
    \includegraphics[width=0.89\textwidth]{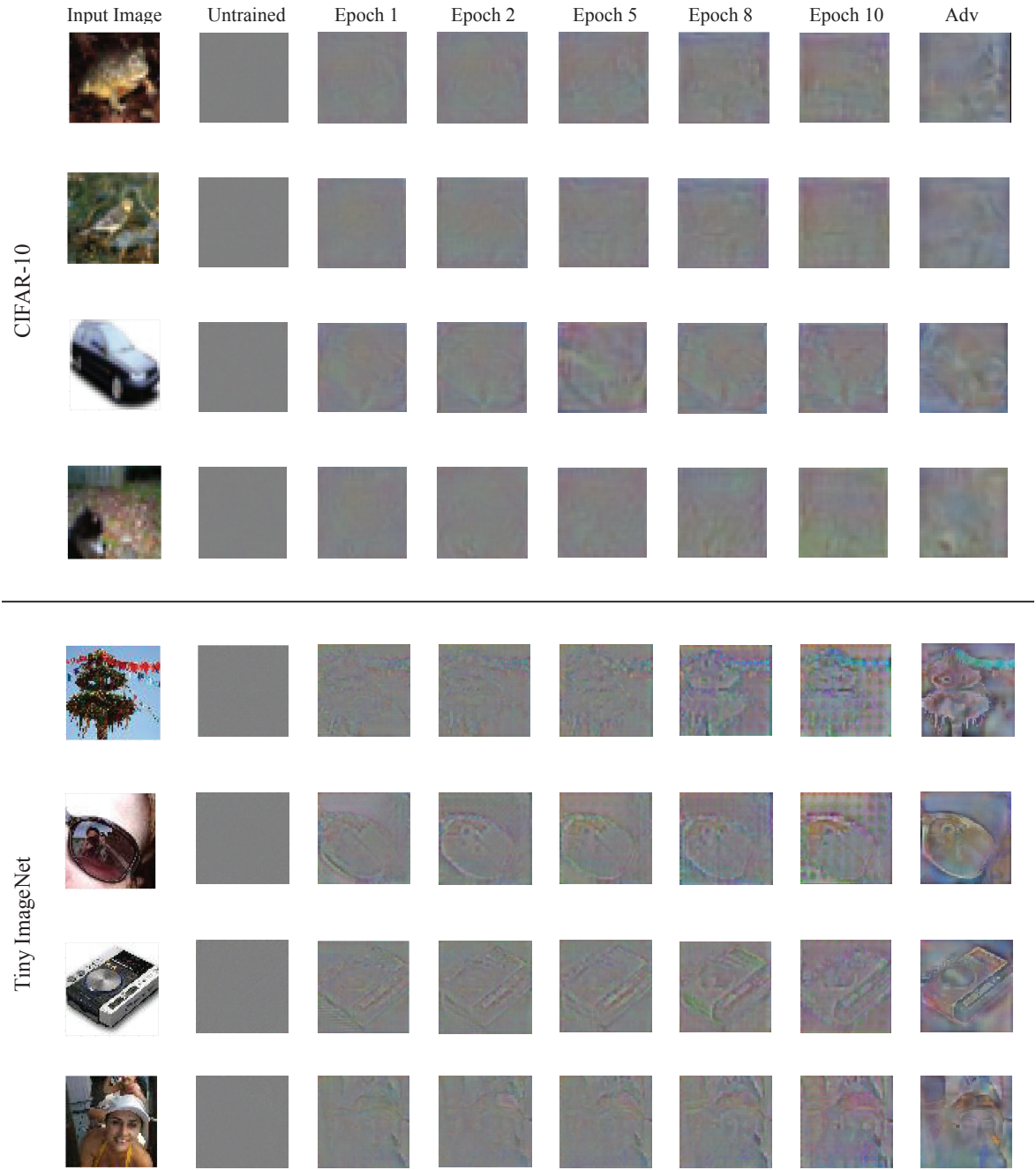}
    \caption{Here we show supplementary examples similar to those in \figref{expected_img} in
    the main text. The top four images are the results on CIFAR-10, while the bottom four
    images are those on Tiny ImageNet. The first column shows the input image $\bm{x}$ in each example.
    The other columns show the images generated by adversarial transformations 
    with the $f_b$ models that are untrained, trained with an increasing number of epochs, and 
    adversarially trained, as labeled on the top line. Each of those images is a visualization of
    the normalized difference $\bm{x}-\E_{\tilde{g}\sim\mathcal{T}}\tilde{g}(\bm{x})$,
    where the expectation is estimated using 5000 samples.
    It is evident that as the number of training epochs increases, the
    expectation over transformation
    $\E_{\tilde{g}\sim\mathcal{T}}\tilde{g}(\bm{x})$ drifts further away from $\bm{x}$,
    and the adversarially trained $f_b$ model produces an even larger difference.
  }\label{fig:exp_app}
\end{figure*}

Figure~\ref{fig:expected_img} in the main text shows a few examples of 
the difference between an input image $\bm{x}$ and its expectation over transformation,
that is, the image of normalized $\bm{x}-\E_{\tilde{g}\sim\mathcal{T}}\tilde{g}(\bm{x})$.
We now provide more samples of 
$\bm{x}-\E_{\tilde{g}\sim\mathcal{T}}\tilde{g}(\bm{x})$ images
on both CIFAR-10 and Tiny ImageNet (see \figref{exp_app}). 

\begin{figure*}[t]
  \centering
  \includegraphics[width=0.84\textwidth]{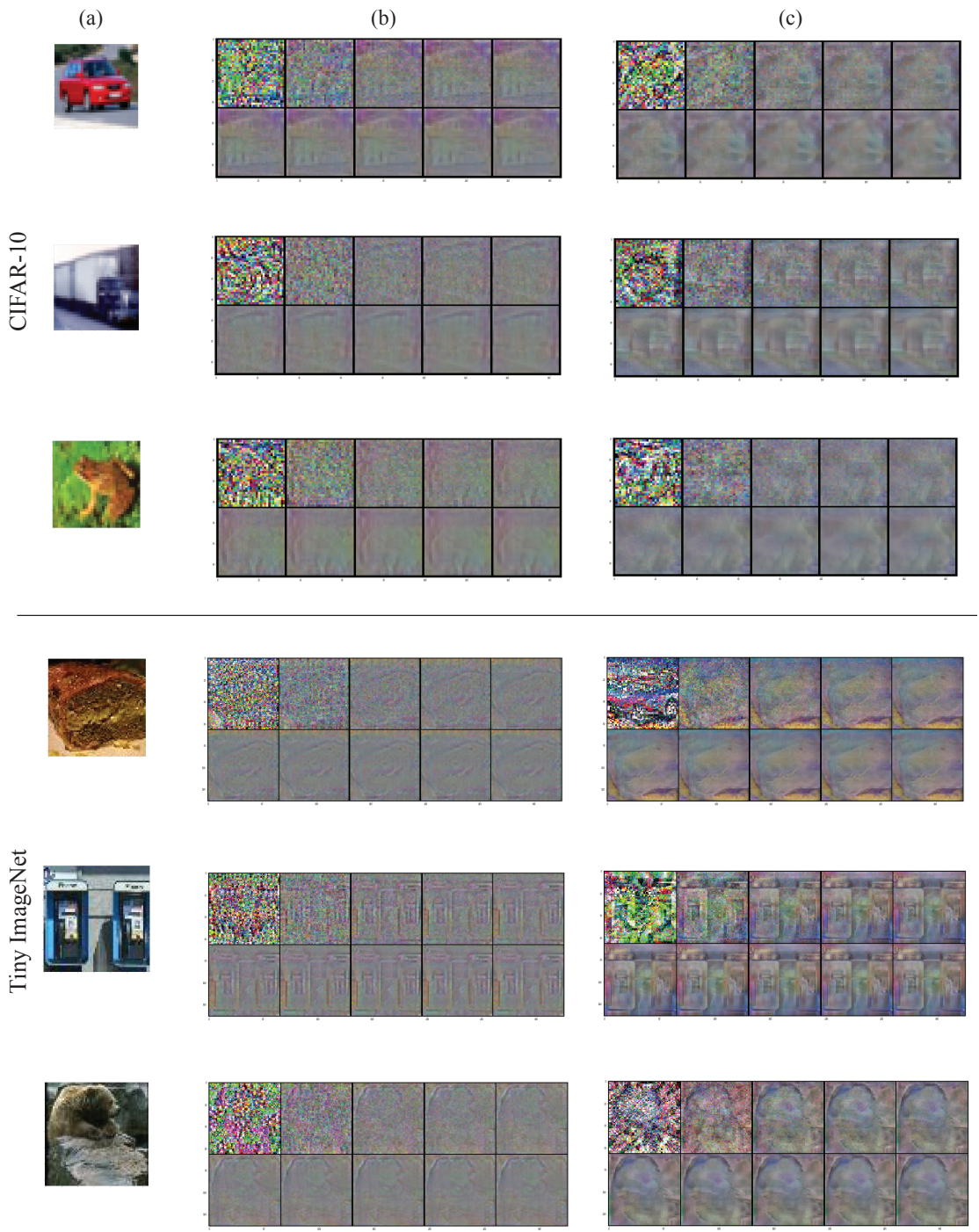}
  \caption{
        Here we visualize the normalized difference between an input image (shown in column (a))
        and its expectation over transformation image $\E_{\tilde{g}\sim\mathcal{T}}\tilde{g}(\bm{x})$
        in our defence model. The top three examples are from CIFAR-10, and the bottom three 
        are from Tiny ImageNet. 
        Column (b) shows the results using $f_b$ models with standard training,
        while column (c) are results with adversarial training.
        The Expectation over transformation in each example is
        estimated using an increasing number of samples. The ten sub-images (from left to right, top to bottom) in each group 
        of column (b) and (c) are results in which the expectations over 
        transformation are estimated using
        1, 10, 50, 100, 200, 500, 1000, 2000, 5000, 10000 samples of $\tilde{g}(\cdot)$, respectively.
  }\label{fig:exp_iter}
  \vspace{-1mm}
\end{figure*}

\textit{Discussion.}\;
In~\cite{tsipras2018robustness}, Tsipras et al. presented an interesting finding.
They visualized the loss gradient with respect to input pixels, and found that if the model 
is adversarially trained, such a loss gradient is significantly \emph{human-aligned}---they 
align well with perceptually relevant features (e.g., see Figure 2 in their paper). But if the model is not adversarially trained, 
the loss gradient appears like random noise. Here, we discover that 
the normalized difference $\bm{x}-\E_{\tilde{g}\sim\mathcal{T}}\tilde{g}(\bm{x})$
is also human-aligned, exhibiting perceptually relevant features, as shown in \figref{exp_iter}.
In contrast to the discovery in~\cite{tsipras2018robustness}, we found that 
$\bm{x}-\E_{\tilde{g}\sim\mathcal{T}}\tilde{g}(\bm{x})$ is always human-aligned. 
Even if the model $f_b$ is not adversarially trained, 
the difference image $\bm{x}-\E_{\tilde{g}\sim\mathcal{T}}\tilde{g}(\bm{x})$ still exhibits
perceptually relevant features, as along 
as they are trained with sufficient number of epochs (see \figref{exp_app}).
If the model $f_b$ is adversarially trained, those perceptually relevant features become more noticeable.


\section{Discussion on Computational Performance}
Our defense demands lower training cost than the standard adversarial training.
For example, on CIFAR-10 dataset, our method takes \textbf{82 minutes}
to train a ResNet18 model for 80 epochs. 
This time cost is close to the standard (non-adversarial) training, which takes 56 minutes
for the same setting. In contrast, the standard adversarial training takes \textbf{460 minutes} for 
the same number of epochs and the same network structure. 
Notice that the lower training cost in our method is obtained without sacrificing its robustness performance. In fact, 
as shown in \tabref{cmp_cifar} in the main text and \tabref{cmp_tiny} here, our defense offers
much stronger robustness.


The inference cost of our defense is more expensive than adversarially trained models,
because the input image $\bm{x}$ during the inference also needs to be transformed by $g(\cdot)$.
In our experiments, our defense takes 17 seconds to predict the labels of 10000 images in CIFAR-10,
while the adversarially trained model and the standard model (without adversarial training) both take 4 seconds.
This is the cost we have to pay in exchange for stronger robustness.
We argue that this is worthy cost to pay because in comparison to network training cost, 
the inference cost is negligible.
In fact, almost all adversarial defense methods that rely on input
transformation~\cite{guo2018countering,song2018pixeldefend,samangouei2018defensegan}
have a performance overhead at inference time.  
For example, PixelDefend~\cite{song2018pixeldefend} projects the input to a pre-trained
PixelCNN-represented manifold through 100 steps of L-BFGS iterations. 
Their transformation is about 10$\times$ slower than ours
even when our method uses the same network structure in $f_b$ as their PixelCNN.



\end{document}